\documentclass[acmsmall,authorversion]{acmart}

\AtBeginDocument{%
  \providecommand\BibTeX{{%
    \normalfont B\kern-0.5em{\scshape i\kern-0.25em b}\kern-0.8em\TeX}}}

\setcopyright{acmlicensed}
\acmJournal{PACMCGIT}
\acmYear{2021} \acmVolume{4} \acmNumber{1} \acmArticle{} \acmMonth{5} \acmPrice{}\acmDOI{10.1145/3451259}

\acmConference[I3D '21]{Symposium on Interactive 3D Graphics and Games}{April 20-22}{2021}
\acmBooktitle{Symposium on Interactive 3D Graphics and Games (I3D '21), April 20-22, 2021}
\acmSubmissionID{12}

\newcommand{\AJ}[1]{{\color{blue}}}

\usepackage{ulem}

\newcommand{\REM}[1]{}
\newcommand{\ADD}[1]{#1}

\usepackage{multirow}
\usepackage{amsmath}

\DeclareMathOperator*{\argmax}{argmax}

\begin{document}

%\title{ Cheating Detection in Competitive Games}
\title{Robust Vision-Based Cheat Detection in Competitive Gaming}
%\title{A Machine-Learning Algorithm for Purely Visual Detection of Cheating in Competitive Video Games}
%\title{Deep Neural Networks Can Detect Cheating in Competitive Games from Only Visual Output}

\author{Aditya Jonnalagadda}
%%\authornote{Authors contributed equally to this research.}
%\email{aditya_jonnalagadda@ucsb.edu}
\affiliation{%
  \institution{University of California, Santa Barbara}
  \city{Santa Barbara}
  \country{} %\country{USA}
  }
  %\state{California}
  %\postcode{93117}

\author{Iuri Frosio}
%%\authornotemark[1]
\affiliation{%
  \institution{NVIDIA}
  \city{Santa Clara}
  \country{} %\country{USA}
}

\author{Seth Schneider}
\affiliation{%
  \institution{NVIDIA}
  \city{Santa Clara}
  \country{} %\country{USA}
}

%%\author{Jan Kautz}
\author{Morgan McGuire}
\affiliation{%
  \institution{NVIDIA}
  \city{Santa Clara}
  \country{} %\country{USA}
}

\author{Joohwan Kim}
%%\authornotemark[1]
\affiliation{%
  \institution{NVIDIA}
  \city{Santa Clara}
  \country{} %\country{USA}
}

\renewcommand{\shortauthors}{Jonnalagadda, Frosio, Schneider, McGuire and Kim}

\begin{abstract}

Game publishers and anti-cheat companies have been unsuccessful in blocking cheating in online gaming.
We propose a novel, vision-based approach that captures the frame buffer's final state and detects illicit overlays. 
To this aim, we train and evaluate a DNN detector on a new dataset, collected using two first-person shooter games and three cheating software.
We study the advantages and disadvantages of different DNN architectures operating on a local or global scale.
We use output confidence analysis to avoid unreliable detections and inform when network retraining is required.
In an ablation study, we show how to use Interval Bound Propagation (IBP) to build a detector that is also resistant to potential adversarial attacks and study IBP's interaction with confidence analysis.
Our results show that robust and effective anti-cheating through machine learning is practically feasible and can be used to guarantee fair play in online gaming.

\end{abstract}





%\begin{comment}

\begin{teaserfigure}
  \centering
  \includegraphics[width=0.9\textwidth]{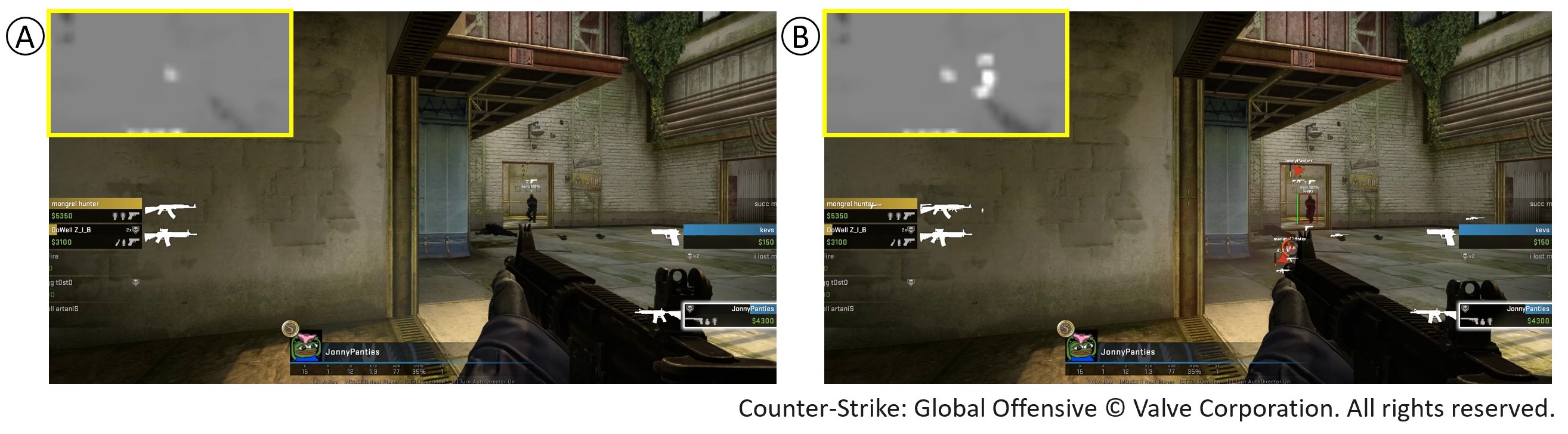}
  \vspace{-1em}
      \caption{Panel A: \textit{Counter-Strike:Global Offensive }\cite{Valve2012} game frame, Panel B: cheating information added to the game frame using AimWare. Our DNN cheating detector outputs: the probability that a frame contains visual hacks (P), the uncertainty level of such estimate (U), and a map with each pixel's probability to belong to a cheating frame (shown as inset). DNN output: Panel A: clean frame with high confidence (P=$0.0026\%$, confidence=$99.7\%$, computed as $(1-\text{U})$), Panel B: cheating frame (P=$99.995\%$ and high confidence=$99.99\%$).}
  %\
  \Description{}
  \label{fig:teaser3}
  \centering
\end{teaserfigure}

%\end{comment}
%\raggedbottom
%\flushbottom
%

\maketitle
\section{Introduction / Motivation}

Cheating is widespread in the quickly growing competitive video gaming field, also called esports\ADD{, among professional athletes and casual gamers alike. \textit{Counter-Strike:Global Offensive} (CS:GO), today's most popular esports shooter game, often sees professional athletes cheat even in official tournaments \cite{tournament_cheat1, tournament_cheat2, tournament_cheat3}}. 
Cheating hacks and software are easily accessible to casual gamers, as many websites offer cheating \REM{SW}\ADD{software} for a small or no fee~\cite{aimware, iniuria}. \AJ{Today's video games typically holds ten or more players in a single match. As a consequence, } Even a small percentage of cheaters can ruin the gaming experience for a large percentage of players: if 6\% of players are cheating, the chance of meeting at least one cheater is 42.7\% in 5 vs. 5 matches (e.g., CS:GO or Valorant) and 49.4\% in 6 vs. 6 matches (e.g., Overwatch).

Two types of cheating strategies are adopted in competitive shooter games:
aim assistance (also known as aimbots) is typically triggered by pressing a hotkey to modify the mouse input and accurately aim at the opponents. Preliminary results show that machine learning can be used to analyze the behavior of natural players and identify cheaters in this case~\cite{Alayed_CheatDetection}. We focus on identifying the second major category of cheating, known as visual hack, wallhack, or extrasensory perception, which renders illicit information (e.g., opponents hiding behind walls, Fig.~\ref{fig:teaser3}) on top of the original game scene and have received less attention so far.

\AJ{Game publishers and anti-cheat companies' efforts to block cheating (and visual hacks in particular) have been only partially successful. }Efforts to block cheating are based on three main approaches: first, automated analysis of game logs through machine learning can detect non-human activities, such as naive aim assistance (e.g., instant aim lock~\cite{Alayed_CheatDetection}), but cheating software is also evolving to make cheating hardly detectable (e.g., human-like aim lock with enhanced but limited accuracy). Second, inspecting the host machine's memory can effectively detect suspicious activities, but it also raises security \ADD{and privacy} concerns.\AJ{ depending on how intrusive the inspection is. At the same time, some cheating software can get through even the most intrusively inspecting anti-cheat protection.} The third approach, i.e., manual inspection of gameplays reported by other players, is so far the most successful\AJ{, especially in catching sophisticated cheaters}, but it is laborious and can ban non-cheating players based on mistaken conclusions.

We introduce and analyze a novel, vision-based, cheat detection procedure for visual hacks.
\ADD{In the battle against cheating, visual detection can be seen not as an alternative, but as a complementary approach to non-visual detection, with some additional advantages: since detection is performed on the last state in memory before displaying, further frame modifications by the cheater are not allowed; furthermore, the visual nature of the task allows for a simple interpretation of the results for a human observer.}
%\ADD{Using a vision-based approach, we are leveraging the fact that visual hacks are bound to modify visual game information to convey the cheating information to the gamer. Thus, by tying ourselves with the gamer, we are forcing the cheat manufacturers into an impossible situation of choosing between escaping detection by our system and providing useful cheating information.}
Our contributions are multifold: first, we show that training a Deep Neural Network (DNN) to detect visual hacks is practically feasible; we compare detection at a global or local scale to define the best DNN architecture.
We show that taking into account the confidence of the DNN output (as in \cite{NIPS2018_7580}) allows avoiding false positives while keeping the cheating detection rate high.
Furthermore, we show how to use the confidence statistics to automatically detect the need for DNN re-training, which may occur every time the cheating software changes.
Lastly, once our solution is in action, it is easy to imagine that cheating software may evolve to get around or attack it. \ADD{In an ablation study, w}\REM{W}e show how Interval Bound Propagation (IBP~\cite{DBLP:journals/corr/abs-1810-12715}) can protect our DNN against practical adversarial attacks \ADD{potentially put in place by cheating companies}\REM{; in an ablation study, we}\ADD{and also} analyze the positive interaction between IBP and the confidence estimate.
A last contribution is represented by the dataset that we captured and used for training and benchmarking our solution, \ADD{publicly available at \href{https://github.com/NVlabs/deep-anticheat}{https://github.com/NVlabs/deep-anticheat}.}\REM{publicly available at~\cite{public_dataset}.}
\section{Prior work}

\subsection{Anti-cheating solutions}

Data mining techniques can examine the memory of the host machine~\cite{Philbert_DataMining} to detect the presence of specific cheating software, but these solutions raise privacy concerns, as the anti-cheating software can access the entire memory of the host.
To prevent a privacy breach, some propose server-side cheat-detection based on game logs to find anomalies in-game events and performance metrics~\cite{Alayed_CheatDetection}. Given the large amount of data in the logs, machine learning techniques like decision trees, naive Bayes, random forests, DNNs, or support vector machines~\cite{6032016}\ADD{\cite{10.1007/978-3-030-04648-4_21}} are a good fit for cheat detection, but currently available anti-cheating systems mainly focus on non-visual features~\cite{FACEIT, Punkbuster, Hestianet, VACNet}\ADD{~\cite{9154512}}. For example, Punkbuster~\cite{Punkbuster} scans the memory of the player machine, HestiaNet~\cite{Hestianet} analyzes data using machine learning, VACNet~\cite{VACNet} analyzes players actions and decisions. 
\REM{The algorithms behind these commercial tools are unpublished and thus unavailable for objective evaluation and comparison, whereas}
\ADD{Since the algorithms and source code of these commercial tools are unpublished, the details are unavailable for objective evaluation and comparison. To the best of our knowledge, }
anti-cheating solutions based on processing of the rendered frame, a much safer approach from the privacy point of view, are currently nonexistent. \ADD{It is essential to notice that this approach is safe from the privacy point of view, as it only requires as input the rendered frame residing in the GPU buffer before showing it on display.
Notice that frames do not need to be transmitted if the detection is performed on the local machine.
Furthermore, the computational overhead is limited, as frame analysis can be performed with low priority and not necessarily in real-time.}

\subsection{Confidence analysis}

\ADD{In the context of visual cheating detection, it is crucial to estimate not only the probability that a given frame contains cheating information but also the confidence of such an estimate.} We leverage confidence analysis to improve two aspects of our cheating detection system that are fundamental for its practical adoption: to prevent reporting \REM{legit}\ADD{legitimate} gamers as cheaters; and to identify the need for DNN retraining\ADD{, which may occur every time a cheating software is modified, or a new one introduced}.

It has been indeed demonstrated that, for a DNN classifier, the output probabilities are poor indicators of its confidence~\cite{NIPS2018_7580}.
Procedures to estimate the DNN confidence have been consequently proposed.
A simple method is based on dropout in the intermediate layers~\cite{2018arXiv181106817M}: by collecting sufficient output statistics for any given input, two metrics (the variation ratio and mutual information) that correlate well with confidence level can be computed. \ADD{However, such a method is computationally demanding, and it is based on intuition more than a well-established theory.}
Taking into consideration the confidence output not only at inference time but also during training provides further advantages.
Sensoy et al. describe a principled approach, where the loss function models prior and posterior multivariate classification distributions with Dirichlet distributions representing not only the target classes of the classifier but also an additional class for uncertain inputs.
This leads to a DNN that outputs the probability that the input belongs to a given class and the level of confidence, improving the ability to detect queries that do not belong to the training distribution and the DNN robustness against adversarial examples~\cite{NIPS2018_7580}.

\subsection{Adversarial robustness}
Adversarial attacks can fool DNNs using images that appear innocuous to humans but produce a target output~\cite{carlini2019evaluating, xu2019adversarial}.
\REM{Such an attack may be used against our DNN cheating detector.}
\ADD{In the context of cheating detection, in addition to adding visual hacks to the rendered frames, cheating software may also add an adversarial pattern to fool our DNN detector.
Although generating an adversarial perturbation that is imperceptible to the cheating gamer is not strictly needed, an attacker should anyway be careful not to corrupt the frame to the point where playing becomes uncomfortable, or the adversarial perturbation reduces the visibility (and thus the utility) of the visual hack.}

Several attacking strategies have been proposed. Some of them, like the Fast Gradient Sign Method (FGSM~\cite{43405}), Projected Gradient Descent (PGD~\cite{Kurakin2017AdversarialEI}), and its improved version with random start (Madry~\cite{DBLP:conf/iclr/MadryMSTV18}) can be computed rather quickly, and thus meet one of the requirements to be potentially employed by a cheating software: the possibility to be computed on the fly and consequently applied to a cheating frame in real-time.
More powerful attacks, like black-box attacks, are not practically feasible due to their computational complexity. 
FGSM and PGD belong to the so-called class of white-box attacks, where the attacker has access to the full DNN model. This may seem an insurmountable issue for an attacker, but researchers have demonstrated that adversarial attacks do transfer, i.e., attacks computed on a given DNN are often successful on a different DNN trained on the same dataset~\cite{DBLP:journals/corr/abs-1809-02861}.
Furthermore, researchers also discovered universal attacks, i.e., perturbations that can be computed offline (thus with zero computation overhead) and applied to an entire set of images, with a discrete attack success rate~\cite{8099500}.

\ADD{Together with attacking methods, the researchers also developed several defense techniques, but a solution that guarantees complete protection on general multi-class problems is not yet available.}
Because of its simplicity, adversarial training has become a popular method to prevent attacks and create robust DNNs~\cite{43405}.
\ADD{It consists of adding adversarial images, generated with fast methods like FGSM, in the training dataset, so that the DNN learns to classify them correctly; however, the protection is not complete, and there is no mathematical guarantee of the achieved level of protection.} A different defense method, IBP~\cite{DBLP:journals/corr/abs-1810-12715}, can be interpreted as a generalized form of adversarial training: instead of generating training adversarial images, an upper bound for the worst-case \ADD{adversarial} perturbation is propagated through the DNN, allowing to establish a bound on the maximum output change, and training performed on the worst-case scenario.
\ADD{IBP has the additional advantage of being capable of explicitly computing the bound and thus providing a guaranteed level of protection for a given magnitude of the input perturbation.}

Existing defense techniques treat all the classes equally, while our application presents a different opportunity: we are only concerned with false negatives generated by adversarial attacks, while the second case (turning a non-cheating frame into a cheating one) has no practical importance \ADD{for the attacker}. Thus, we consider the case of asymmetric protection for the first time to the best of our knowledge. Similarly, and again for the first time to the best of our knowledge, we also study the interaction of the uncertainty loss~\cite{NIPS2018_7580} and IBP, which may be of interest beyond its application for anti-cheating. 
\section{Dataset}
To train and benchmark our DNN detector, we created a dataset of frames from two widely diffused shooter games, which we call Game1 and Game2 in this manuscript, that are played competitively and have actively running professional tournaments, but for which cheating has not been successfully blocked yet despite the on-going effort by the publishers~\cite{tournament_cheat1, tournament_cheat2, tournament_cheat3}.
Two of the authors played competitive matches in five Game1 maps and four Game2 maps and stored the game\ADD{-}play data using the in-game match saving features\ADD{, which saves rendering resources and game event history of the given match. When replaying the recorded matches, the game engine re-renders the entire game by using the saved data}. Since the rendering mechanism is the same while playing and replaying the game, the cheating software reacts to replays as if it would do during a real game\ADD{-}play --- we leverage this feature to add \ADD{realistic} cheating overlays in our dataset.

To acquire sets of cheating / non-cheating frames at 60Hz, we used the Open Broadcaster Software (OBS)~\cite{obs}.
We leveraged  OBS' two levels of scene capture to capture clean and cheating images at the same time, as the Game Capture renders the game scene only with no cheating, while the Screen Capture saves the final state of the frame buffer, which is the same shown on display and thus includes the added visual hack~\cite{stevewhims}.
\ADD{Notice that a similar mechanism can be employed to capture the frame buffer's final state once the DNN detector is deployed on the field.
If a screen dump from the game monitor is preferred for practical implementation, some post-processing steps would be required before using it as input to our network.
Firstly, a simple task of identifying the rendered window and, secondly, a more challenging task of alienating the game from overlays would be required.}

\REM{We used widely used cheating software (AimWare~\cite{aimware}, Iniuria~\cite{iniuria} for CS:GO, and Artemis~\cite{skycheats} for Overwatch) to add visual hacks.}
\ADD{We used AimWare~\cite{aimware}, Iniuria~\cite{iniuria} for Game1, and Artemis~\cite{skycheats} for Game2, which are among the most widely used cheating software on these games, to add visual hacks, thus leveraging actual cheating information in the wild for training and testing the network.}
For testing, benchmarking, and analysis, we divided the dataset into three sets $S_A$, $S_B$, and $S_C$, each containing a different number of training, validation, and testing frames, as well as maps (see Table~\ref{tab:dataset}).
$S_A$ contains data from a single game map; its variability is limited compared to $S_B$ and $S_C$.
The union of $S_A$ and $S_B$ allows doing experiments in the case of different cheating software used for the same game: it allows testing the hypothesis that a DNN detector trained on a given cheating software does generalize to another one, which is likely to add different cheating features to the game image.
\ADD{We did not include the case of the same cheating software on different games, as cheating software are generally game specific.
The entire dataset allows performing experiments in which a DNN detector is trained for different cheating software and games, and thus allows us to verify if a single DNN can be deployed in a practical scenario or whether multiple models are needed.}

\ADD{Keeping in mind that different gamers may configure the cheating software differently accordingly to their preference or, in the worst-case scenario for the DNN detector, to keep the minimum cheating information while trying to fool it, in our dataset we}
\REM{We} also controlled the amount of visual hacks to vary the visual impact and corresponding utility for the cheaters.
\ADD{Experienced players might differ from amateur players in using less cheating information, making it essential for the network to be robust against the level of cheating information present in the frame. Experienced players might also be able to access diverse game scenarios, which can be seen as a useful feature for training data, but also demands good scenario generalization capabilities of the DNN.}
For the sets $S_{B}$ and $S_{C}$ we defined (somehow arbitrarily) three settings as follows: a \emph{minimal} setting, where cheating information only indicated opponent position with a small pixel footprint; a \emph{full} setting, that entirely change opponent appearance with additional augmentations on illicit information such as health points and weapon types; and a \emph{medium} setting which stays approximately in the middle between these two.
To facilitate further research, development and benchmarking, the dataset from Game1 is made publicly available \ADD{at \href{https://github.com/NVlabs/deep-anticheat}{https://github.com/NVlabs/deep-anticheat}}\REM{ at \textit{hidden URL}}.

\begin{table*}
  \caption{Characteristics of the recorded dataset including two cheating software, three games, and a number of maps played in competitive matches; the dataset is divided into three sets $S_{Z}, Z = A, B, C$; each set include training, validation, and test frames:  $S_{Z} = {S_{Z}^{train} \bigcup S_{Z}^{val} \bigcup S_{Z}^{test}}$. The cheating software configuration indicates the amount of cheating information.} %rendered into the frames.}
  \vspace{-1em}
  \footnotesize{
  \resizebox{\columnwidth}{!}{
  \begin{tabular}{ccccccccl}
    \toprule
    $Set$ & Game & Cheating software & Cheating software config. & Train ($S_{Z}^{train}$) & Validation ($S_{Z}^{val}$) & Test ($S_{Z}^{test}$) & Number of Maps\\
    \midrule
    %S_{CS:GO}^{Cheat 1}}
    $S_{A}$ &  Game1 & AimWare  & full & 480 frames & 340 frames & 500 frames & 1\\
    $S_{B}$ & Game1 & Iniuria & full, medium, minimal & 861 frames & 360 frames & 1020 frames & 4\\
    $S_{C}$& Game2 & Artemis & full, medium, minimal & 1320 frames & 760 frames & 1460 frames & 4\\
  \bottomrule
\end{tabular}}}
\label{tab:dataset}
\end{table*}

\section{Network design}
Our DNN detector takes in input a full resolution, $1080\times1920\times3$ RGB frame. Although computationally demanding, processing at full resolution is required to guarantee that tiny cheating details are passed to the DNN.
A sequence of 4 convolutional layers processes the frame, each followed by leaky ReLUs\ADD{~\cite{Nair2010RectifiedLU}}, containing 48 $5\times5$, 48 $5\times5$, 32 $3\times3$, and 16 $3\times3$ filters with stride 3, 3, 2, and 2, respectively.

The DNN detector has two heads used for local and global detection, originating from the fourth convolutional layer.
The global head aims to output the probability that the input frame has been altered by visual hack.
The global head has a reduction layer followed by a fully connected layer ($16\times2$) and a softmax layer. To identify the most effective architecture for the global head, we considered three different reduction layers: a fully connected layer, which adds parameters to the DNN and thus potentially leads to better results, but has the drawback of making the detection dependent on the position of the cheating features within the image; a max-pooling layer, that does not suffer from the same drawback, but whose output may be affected by outliers, making it also prone to adversarial attacks; and an average pooling layer, which is also spatially invariant, although the average operation may dilute evidence of a visual hack.
To train the global head, we minimize the traditional binary cross-entropy:
\begin{equation}
    CE_{global} = -(1/N) \sum_i y_i log(p(y_i)) + (1-y_i) log(1-p(y_i)),
\end{equation}
where $y_i=1$ if the i-th frame in the training dataset with $N$ frames is a cheating one, and zero otherwise, and $p(y_i)$ is the corresponding DNN output. \ADD{Notice that training does not require cheating / non-cheating frame pairs.}

The local detection head includes one additional convolutional layer with a unique $1\times1$ filter followed by a logsig\ADD{~\cite{logsig}} activation.
Each pixel in output (insets in Fig.~\ref{fig:teaser3}) ideally represents the probability that the corresponding input area belongs to a cheating image.
The reason for introducing the additional local head detector is the following: during training, a global detector may easily concentrate its attention on the most evident clue in the scene while forgetting minor or poorly evident cheating information. On the other hand, a local detector is forced to focus its attention on any local cheating feature, neglecting the rest of the image content. Therefore we speculate that the gradient coming from the local head may act as an implicit attention mechanism and force the global classifier to focus on all the visual hack features present in the scene.
The local detector also has an additional advantage in terms of explainability, as the output map is easy to interpret and allows the visual identification of the hacks in the frame. In contrast, the global detector does not provide this kind of spatial information unless one resorts to saliency analysis algorithms like Grad-Cam~\cite{8237336}, which is, on the other hand, time-consuming, noisy, and hard to interpret.

A peculiar aspect of training the local head in our system is that we do not have access to the ground truth at the pixel level, as we only know if a frame contains cheating information or not, but we do not know anything about its spatial localization.
Nonetheless, we discovered that training is feasible even providing as ground truth to each pixel the frame label: after training, any pixel that does not contain cheating information will have a $50\%$ probability of coming from a cheating frame, as the local detector has no local clues, while pixels containing cheating information will be characterized by a probability larger than  $50\%$.
To train the local head, we minimize the binary cross-entropy again:
\begin{equation}
\begin{split} 
    CE_{local} &= -(1/N) \sum_i \frac{1}{W*H} \sum_{m,n} y_i^{smooth} log(x_{i,m,n}) \\
    &+ (1-y_i^{smooth}) log(1-x_{i,m,n}),
\end{split}
\end{equation}
where $y_i^{smooth}$ is the smoothed label of the i-th frame in the training dataset (i.e., $y_i^{smooth}=0.9$ for a cheating frame, 0.1 otherwise), $x_{i,m,n}$ is the output of the local detector in position $m,n$, whereas $W$ and $H$ are the width and height of the output map.
The full DNN detector is then trained using a combination of the local and global loss:

\begin{equation}
    CE_{combined} = 0.5 (CE_{global} + CE_{local}).
\end{equation}.

For training on any of the sets $S_Z, Z=\{A, B, C\}$ or their union, we use the Adam optimizer\ADD{~\cite{Adam_optimizer}}, initial learning rate of $1e-4$, with a decay of $0.1$
%( patience of $20$ epochs), 
and a total of 1000 epochs. 
%(early stopping patience of $125$ epochs).

\subsection{Results}

To identify the best reduction layer in the global head and test the effect of the additional local head onto the global detection performances, we trained our DNN detector using the fully connected, max pooling, or average pooling layer in the global head, minimizing $CE_{global}$ or $CE_{combined}$ on data from $S_{A}^{train} \bigcup S_{B}^{train} \bigcup S_{C}^{train}$, and measure the accuracy of the DNN global head on each testing set separately.
Data in Table~\ref{tab:network-design_a} show that the detector's overall accuracy generally improves when moving from the fully connected to the max-pooling to the average pooling layer.
Furthermore, when average pooling is adopted in the global head, the accuracy increases when moving from minimization of the global cost function, $CE_{global}$, to the combined minimization of the global and local detection error.
Overall, this table suggests that the highest accuracy is achieved using an average pooling layer in the global head and combined training of the local and global detector heads.

To better highlight the advantages and disadvantages of adopting a different reduction layer in the global head and the two cost functions, we computed the accuracy of the DNN detector, trained using data from the entire dataset, and tested on $S_B^{test}$ and $S_C^{test}$, 
measuring the accuracy as a function of the amount of cheating information present in the frames (full, medium, or minimum).
The results in Table~\ref{tab:freq_various_3} show that, even when disentangling the cheating information level, the use of the combined cost function together with the average pooling layer in the global head leads to the overall better accuracy, thus confirming the previous results.
This same Table highlights that, when a minimum amount of cheating information is added to the scene, the detection task becomes more challenging, as expected, but at the same time also the cheating information becomes less useful for the cheater because of the tiny size, low contrast and visibility of the added visual hack. 
Surprisingly, we also found that the accuracy is higher for a medium level of cheating information than the full cheating case.
We speculate that the full cheating configuration may add cheating information that is possibly tiny, poorly relevant, or not always present in any frame, thus hard to be learned by the DNN detector that eventually treat it as noise, thus decreasing the overall accuracy.
More investigation is needed to better clarify this aspect, which, however, does not significantly affect the performance achieved overall by the DNN detector.

\begin{table*}
  \caption{\REM{Accuracy of the DNN detector, trained on $S_{A}^{train} \bigcup S_{B}^{train} \bigcup S_{C}^{train}$,  for different reduction layers in the global head and different training losses. The best accuracy for each set is highlighted in bold.}\ADD{Accuracy of the DNN detector, trained on $S_{A}^{train} \bigcup S_{B}^{train} \bigcup S_{C}^{train}$,  for different reduction layers (fully connected, max pooling, average pooling) in the global head and different training losses ($CE_{global}$, $CE_{combined}$). The best accuracy for each set is highlighted in bold.}}
  \label{tab:network-design_a}
  \vspace{-1em}
  \footnotesize{
  \resizebox{\columnwidth}{!}{
  \begin{tabular}{ccccc|cccc}
    \toprule
    & \multicolumn{4}{c}{$CE_{global}$} & \multicolumn{4}{c}{$CE_{combined}$} \\ \cmidrule(r){2-5} \cmidrule(r){6-9}
    Global head reduction layer & $S_{A}^{test}$ & $S_{B}^{test}$ & $S_{C}^{test}$ & $S_{A}^{test} \bigcup S_{B}^{test}  \bigcup S_{C}^{test}$ & $S_{A}^{test}$ & $S_{B}^{test}$ & $S_{C}^{test}$ & $S_{A}^{test} \bigcup S_{B}^{test}  \bigcup S_{C}^{test}$\\
    \midrule
    fully connected & 0.97 & 0.76 & 0.64 & 0.74 & 0.97 & 0.75 & 0.59 & 0.71\\
    max pooling & 1.00 & 0.7 & \textbf{0.9} & 0.85 &0.99 & 0.71 & 0.83 & 0.82\\
    average pooling & 1.00 & 0.82 & 0.85 & 0.87 & \textbf{1.00} & \textbf{0.86} & 0.88 & \textbf{0.89}\\
  \bottomrule
\end{tabular}}}
\end{table*}
% weights for test set (0.17, 0.34, 0.49)

\begin{table*}
  \caption{\REM{Table~\ref{tab:network-design_a} repeated for full, medium, or minimal configurations of the cheating software.}\ADD{Accuracy of the DNN detector, trained on $S_{A}^{train}+S_{B}^{train}+S_{C}^{train}$, for different reduction layers (fully connected, max pooling, average pooling) in the global head and different training losses ($CE_{global}$, $CE_{combined}$), and tested on $S_{B \bigcup C}^{test} = S_B^{test} \bigcup S_C^{test}$, for different configurations of the cheating software generating the full, medium, or minimal cheating information. The best accuracy for each set is highlighted in bold.}}
  \label{tab:freq_various_3}
  \vspace{-1em}
  \footnotesize{
  \resizebox{\columnwidth}{!}{
  \begin{tabular}{ccccccl}
    \toprule
    & \multicolumn{3}{c}{$CE_{global}$} & \multicolumn{3}{c}{$CE_{combined}$} \\ \cmidrule(r){2-4} \cmidrule(r){5-7}
    Global head reduction layer & $S_{B \bigcup C}^{test, full}$ & $S_{B \bigcup C}^{test, medium}$ & $S_{B \bigcup C}^{test, minimal}$ & $S_{B \bigcup C}^{test, full}$ & $S_{B \bigcup C}^{test, medium}$ & $S_{B \bigcup C}^{test, minimal}$\\
    \midrule
    fully connected & 0.65 & 0.64 & 0.55 & 0.61 & 0.62 & 0.54\\
    max pooling & 0.81 & 0.86 & 0.67 & 0.8 & 0.86 & 0.67\\
    average pooling & 0.79 & 0.83 & 0.64 & \textbf{0.84} & \textbf{0.88} & \textbf{0.67}\\
  \bottomrule
\end{tabular}}}
\label{network-design_D2}
\end{table*}
\section{Network confidence}
Given that we expect the cheating software to evolve quickly, especially once engaged in the cheating anti-cheating battle, we need our DNN to adapt to new conditions as the ones generated, for instance, by the introduction of a new cheating software.
Given the small size and training time of our DNN detector, re-training (even from scratch) does not constitute a significant obstacle to its adoption.
On the other hand, automatically detecting when DNN re-training is needed, once the detector has been deployed on the field, appears to be more critical.
To handle this problem, we propose resorting to confidence analysis, which in short, \REM{maybe}\ADD{may be} explained as the possibility of raising a red flag when the average confidence of the DNN on the field is low, indicating that the input data are far from the training distribution and therefore re-training is required. 
At the same time, we propose to use confidence also to decrease the probability of reporting legit gamers as cheaters because of misclassification errors\AJ{, which turns out to be a strong constraint for a practical detection system that does not have to affect the quality of the gaming experience for legit gamers}.
We first introduce two existing methods to estimate the DNN confidence and then show how to employ them in cheating detection effectively.

\subsection{Dropout based network confidence}
Michelmore et al.~\cite{2018arXiv181106817M} describe a method based on dropout to estimate a DNN output's confidence. The rationale behind the method is that, even in the presence of noise, the network output should not change much if the confidence is high, and vice-versa.  Following~\cite{2018arXiv181106817M}, we introduce a dropout component (with $p=0.15$) after the average pooling layer of the global head detector and process the same input image $T=64$ times to collect sufficient output statistics.
Among the three metrics proposed in~\cite{2018arXiv181106817M} that correlates with the confidence of the output, we experimentally found \ADD{the }best results when using the Variation Ratio, $VR[x]$, which expresses the output dispersion as:

\begin{comment}
\begin{equation}
  \begin{split}
  &\text{VR[x]} = 1 - \frac{f_{x}}{T}, \\
  &\text{ where, }c^{*} = \argmax_{c=1,2} \sum_{t=0}^{T-1}1[y^{t}=c] \text{ and } f_{x} = \sum_{t=0}^{T-1}1[y^{t}=c^{*}]
  \end{split}
\end{equation}
\end{comment}

\begin{equation}
  \begin{split}
  \text{VR[x]} = 1 - \frac{f_{x}}{T}, & \;\; \text{where,} \;\; c^{*} = \argmax_{c=1,2} \sum_{t=0}^{T-1}1[y^{t}=c] \\
  & \;\; \text{ and } \;\;\;\; f_{x} = \sum_{t=0}^{T-1}1[y^{t}=c^{*}]
  \end{split},
\end{equation}
where $x$ is the input, $y^{t}$ is the predicted class for the pass $t$, $c^{*}$ is the most frequently predicted class in $T$ passes.
When confidence of the DNN is high, $VR[x]$ approaches zero, whereas it grows otherwise (to a maximum value of 0.5 for the two classes problem associated with our DNN detector).

\subsection{Uncertainty based network confidence}

\begin{table}
  \caption{\REM{Effect of the adoption of Variation Ratio VR[x]~\cite{2018arXiv181106817M} and uncertainty loss~\cite{NIPS2018_7580} on TP, FN, FP, and TN, generated by our DNN detector. The Non-confident section reports the frames in each class that are discharged because $VR[x]>0$.}\ADD{Effect of the adoption of Variation Ratio VR[x]~\cite{2018arXiv181106817M} and uncertainty loss~\cite{NIPS2018_7580} on the number of true positives (TP, cheating frames reported as cheating), false negative (FN, missed cheating frames), false positives (FP, non-cheating frames incorrectly reported as containing cheating information) and true negative (TN), generated by our DNN detector. The Non-confident section reports the frames in each class that are discharged because $VR[x]>0$.}}
  \label{tab:reducing_FP}
  \vspace{-1em}
  \footnotesize{
  \resizebox{0.8\columnwidth}{!}{
  \begin{tabular}{cccccccccccl}
    \toprule
    & & \multicolumn{4}{c}{\textbf{Confident}} & \multicolumn{4}{c}{\textbf{Non-confident}}  \\ \cmidrule(r){3-6} \cmidrule(r){7-10} 
    Loss function & Variation Ratio & TP & FN & FP & TN & TP & FN & FP & TN\\
    \midrule
    $MSE_{combined}$ & False & 434 & 1056 & 96  & 1394 & - & - & - & -\\
    $MSE_{combined}$ & True & 392 & 1028 & 51 & 1353 & 42 & 28 & 45 & 41\\
    \midrule
    $UL_{combined}$ & False & 670 & 820 & 2 & 1488 & - & - & - & -\\
    $UL_{combined}$ & True & 519 & 752 & 0 & 1457 & 151 & 68 & 2 & 31\\
  \bottomrule
\end{tabular}}}
\label{with-vr_10}
\end{table}

\REM{Differently from}\ADD{Compared to} the case of $VR[x]$, where the DNN confidence is computed on the fly for any input $x$, Sensoy et al.~\cite{NIPS2018_7580} introduce a more principled approach based on uncertainty loss that can be adopted not only to estimate the confidence at inference time, but also to improve training.
Beyond estimating the output confidence, their approach leads to accuracy improvement and generates more regular decision boundaries that turn out to be useful also to defend against adversarial attacks (see next section).
To estimate the confidence level, the authors of~\cite{NIPS2018_7580} introduce the concept of \emph{evidence} of each output class.
Given the logit vector of our DNN detector, $\{l_0, l_1\}$, we define the evidences $\{e_i\}_{i=0,1}$ and alpha factors $\{\alpha_i\}_{i=0,1}$ as:
\begin{eqnarray}
    e_i = e^{l_i}, & \alpha_i = 1 + e_i,
\end{eqnarray}
and derive the belief for each class, $\{b_i\}_{i=0,1}$ and the total uncertainty $u$ as:
\begin{eqnarray}
    b_i = \frac{e_i}{\sum_j \alpha_j}, \;\;\;\; u = \frac{2}{\sum_j \alpha_j}.
\end{eqnarray}
There is a clear connection between the belief and the probabilities of each class estimated by a traditional softmax layer.
In this framework an additional uncertainty class $u$ is added to represent the probability that the classification is indeed uncertain (notice that $\sum_i b_i + u =1$).
Training of the DNN detector is performed by minimizing the uncertainty loss introduced in~\cite{NIPS2018_7580} for the global head detector.
Among the three losses proposed in the paper, we found that the one based on MSE error (Eq. (5) in~\cite{NIPS2018_7580}) produces more stable results, and therefore we adopted it.
For a detailed description of the cost function, we remind the interested reader to the original paper.
In the following, we will indicate this uncertainty loss as $UL_{global}$.

When training the local and global heads together, to take advantage of the implicit spatial attention mechanism generated by the local head, it is necessary to guarantee that the gradient coming from the two heads have similar nature and magnitude. Otherwise, convergence may be badly affected. For this reason, when using $UL_{global}$ to train the global head, we use a Mean Square Error (MSE) loss (indicated as $MSE_{local}$) to train the local head instead of $CE_{local}$.
The final cost function adopted to train the DNN detector with the uncertainty loss approach is then:
\begin{equation}
    UL_{combined} = UL_{global} + MSE_{local}.
\end{equation}
To estimate a representative baseline using the MSE loss, we also introduce the following loss:
\begin{math}
    MSE_{combined} = MSE_{global} + MSE_{local},
\end{math}
where both the local and global heads are trained to minimize an MSE error with respect to the ground truth labels of the input frames.

\subsection{Avoiding false positives}% through confidence analysis}
The first application we envisioned for confidence analysis is to limit the probability of reporting a legit gamer as a cheater: in few words, a cheating frame is reported if and only if the confidence level is high.
When no confidence criterion is adopted to detect cheating frames, i.e., when the DNN is trained using the $MSE_{combined}$ loss and confidence of the output is not estimated (first row of Table~\ref{tab:reducing_FP}), approximately one-third of the cheating frames are correctly identified as containing cheating information (434 TP over a total of 434 TP + 1056 FN cheating frames); however, 96 FP non-cheating frames are incorrectly classified as cheating, which may lead to an unpleasant gaming experience for legit gamers, incorrectly reported as cheaters.
The second row of the same Table shows the partial benefits of adopting the $VR[x]=0$ as a selection criterion (i.e., we discharge any frame whose variation ratio is not optimal): FP decreases to 51, which improves but does not solve the issue.
The use of the uncertainty loss during training (third row) does provide a more significant improvement, as TP grows to 670, whereas at the same time FP is reduced to only 2, thus suggesting the uncertainty loss training does not only allows to estimate the uncertainty of the DNN output, but also lead to a much more robust classifier.
It is then legitimate to ask how the VR criterion and the uncertainty loss interact: the fourth row in the Table demonstrates that these two criterion\ADD{s} provides somehow complementary information about the confidence of the DNN output. When combined, one pays the cost of slightly reducing the number of TP (down to 519), while, on the other hand, reducing the FP to the optimal value of 0.

Overall, Table~\ref{tab:reducing_FP} demonstrates that confidence analysis is needed to create a detection system that does not suffer from the issue of incorrectly reporting legit cheaters and that the combination of $VR[x]$ and uncertainty loss represents a valid solution to this problem, at least on our dataset. 

\subsection{Detecting the need for re-training}% through confidence analysis}
Re-training of the DNN detector is required anytime a change in the game or cheat features renders its output unreliable:
\AJ{In practice, since we cannot know when new cheating software (or new cheating features) are released,} we need a system that is capable of self-evaluation.
To this scope, we extend the ideas presented in~\cite{NIPS2018_7580} to create a simple statistical tool that is experimentally capable of executing such a task on our dataset.
More in detail, given the alpha factors $\{\alpha_i\}_{i=0,1}$,
we define the Likelihood Ratio ($LR$) as:
\begin{equation}
    LR = max(\alpha_0, \alpha_1) / min(\alpha_0, \alpha_1).
\end{equation}
$LR$ is always greater than or equal to one, and it indicates how likely a sample is to belong to the dominant class. It is also strictly correlated to the DNN confidence: as a matter of fact, a large evidence $e_i$ for one class leads to a large $\alpha_i$ value, small uncertainty $u$, and consequently large confidence and large LR.
In other words, a highly confident DNN detector is expected to have a high LR, whereas, in case of degradation of its performances, LR decreases.

We leverage this simple and intuitive rule to automatically determine the need for re-training as follows.
On the training dataset, we define three average likelihood ratios computed over all the instances of the predicted positive ($LR_{P}^{tra}$), negative ($LR_{N}^{tra}$), and all ($LR_{T}^{tra}$) classes.
These are three target baselines for a DNN detector that operates with performances similar to those measured in training.
Once deployed on the field, we measure the corresponding average likelihood ratios on the field: $LR_{P}^{fie}$, $LR_{N}^{fie}$, and $LR_{Tl}^{fie}$.
We hypothesize that a significant degradation in any of the three likelihood ratios can be interpreted as a signal that DNN re-training is required.

To experimentally test the validity of such hypothesis, we trained four DNN detectors using the $UL_{combined}$ loss function and data either from $S_A^{train}$, $S_B^{train}$, $S_C^{train}$, or $S_A^{train} \bigcup S_B^{train} \bigcup S_C^{train}$.
Then, we tested each detector on each set and measured the ratios between the average training LRs and those achieved on the field.
Beyond reporting these ratios for each training-testing set pair, Table~\ref{red-flag} reports the accuracy, sensitivity, and precision of each DNN on the testing sets.
The sensitivity, defined as the $TP/(TP+FN)$ ratio, describes the percentage of cheating frames that the DNN detector identifies correctly.
Ideally, we want the sensitivity to be close to one, but in the practical context of anti-cheating, we need it to be larger than a minimum value to guarantee that the detector indeed captures at least some of the cheating frames.
The precision, defined as the $TP/(TP+FN)$ ratio, describes how many of the identified cheating frames do indeed contain some cheating information.
In the context of anti-cheating, it is critical to keep the precision as close as possible to one. Otherwise, legit players risk being reported as cheaters.
The Table shows a strong correlation between the decrease in the DNN performance quantified by a decrease in one of the likelihood ratios (as measured without resorting to any ground truth by the DNN on the field) and a degradation in the DNN sensitivity and precision. 
By adopting a simple, empirical criterion suggesting that re-training is needed if at least one of the field likelihood ratios is degraded by $50\%$ or more with respect to its training value, we can guarantee a minimum network precision of 0.9 on $S_C$, which is the most difficult among the three sets, and 0.96 in the case of training on the entire dataset. 
Interestingly enough, this simple criterion suggests that the DNN needs re-training anytime it is trained on a given set and tested on a different distribution, which appears reasonable and aligned with our expectations.
It is important to notice that the re-training criterion reported here does not want to represent an optimal rule, as different thresholds on the LR degradation may be easily set to guarantee even higher performances.
\ADD{The network's ability to detect when re-training is required will also help seamlessly deploy the network in the wild.}

\begin{table*}
  \caption{\REM{Significant decrease of any LR (equal to $50\%$ here) suggests that the DNN detector's performance is sub-optimal, and re-training is required. For each row, the minimum of      $\frac{LR_{tot}^{tra}}{LR_{tot}^{fie}}$, $\frac{LR_{pos}^{tra}}{LR_{pos}^{fie}}$ and  $\frac{LR_{neg}^{tra}}{LR_{neg}^{fie}}$ is highlighted in bold.}\ADD{For four DNN detectors trained on a given frameset and tested on the field on other framesets, we report the ratios between the average training likelihood ratios and those measured on the field. A significant decrease of any LR (equal to $50\%$ here) suggests that the DNN detector's performance is sub-optimal, and re-training is required. For each row, the minimum of      $\frac{LR_{tot}^{tra}}{LR_{tot}^{fie}}$, $\frac{LR_{pos}^{tra}}{LR_{pos}^{fie}}$ and  $\frac{LR_{neg}^{tra}}{LR_{neg}^{fie}}$ is highlighted in bold. The corresponding sensitivity, precision, and accuracy of each DNN, measured on the same test sets, are also reported.}}
  \label{tab:freq}
  \vspace{-1em}
  %\scriptsize{ %
  \footnotesize{
  \resizebox{\columnwidth}{!}{
  \begin{tabular}{ccccccccl}
    \toprule
    Training set & Test set & $\frac{LR_{tot}^{tra}}{LR_{tot}^{fie}}$ & $\frac{LR_{pos}^{tra}}{LR_{pos}^{fie}}$ & $\frac{LR_{neg}^{tra}}{LR_{neg}^{fie}}$
    & Sensitivity = $\frac{TP}{TP+FN}$& Precision = $\frac{TP}{TP+FP}$ & Accuracy & Re-train?\\
    \midrule
    \multirow{3}{*}{$S_{A}^{train}$} & $S_{A}^{test}$ & \textbf{1.04} & \text{1.04} & 1.16 %& 1.04
    & 0.98 & 1 & 0.99 & No \\
    & $S_{B}^{test}$ & \textbf{0.14} & 0.74 & 1.15 %& 0.14
    & 0.17 & 0.99 & 0.58 & \textcolor{red}{Yes} \\
    & $S_{C}^{test}$ & 0.24 & 0.48 & \textbf{0.19} %& 0.19
    & 0.29 & 0.59 & 0.54 & \textcolor{red}{Yes} \\
    \midrule
    \multirow{3}{*}{$S_{B}^{train}$} & $S_{A}^{test}$ & 0.90 & \textbf{0.42} & 7.61 %& 0.42
    & 0.98 & 0.64 & 0.72 & \textcolor{red}{Yes} \\
    & $S_{B}^{test}$ & \textbf{1.28} & 1.44 & 1.78 %& 1.28
    & 0.65 & 0. 97 & 0.81 & No \\
    & $S_{C}^{test}$ & 7.72 & 2.97 & \textbf{0.47} %& 0.47
    & 1 & 0.77 & 0.50 & \textcolor{red}{Yes} \\
    \midrule
    \multirow{3}{*}{$S_{C}^{train}$} & $S_{A}^{test}$ & 0.91 & \textbf{0.33} & 0.93 %& 0.33
    & 1 & 0.51 & 0.52 & \textcolor{red}{Yes} \\
    & $S_{B}^{test}$ & 0.43 & \textbf{0.36} & 0.98 %& 0.36
    & 0.71 & 0.92 & 0.82 & \textcolor{red}{Yes} \\
    & $S_{C}^{test}$ & \textbf{0.81} & 0.95 & 1.0 %& 0.81
    & 0.52 & 0.9 & 0.72 & No \\
    \midrule
    \multirow{3}{*}{$S_{A}^{train} \bigcup S_{B}^{train} \bigcup S_{C}^{train}$} & $S_{A}^{test}$ & 1.87 & 1.41 & \textbf{0.89} %& 0.89
    & 1 & 1 & 1 & No \\
    & $S_{B}^{test}$ & \textbf{0.58} & 0.62 & 1.09 %& 0.58
    & 0.65 & 1 & 0.82 & No \\
    & $S_{C}^{test}$ & \textbf{0.52} & 0.76 & 1.01 %& 0.52
    & 0.45 & 0.96 & 0.72 & No \\
    %1 & 2 &  &  &  &  &  \\
  \bottomrule
\end{tabular}}}
\label{red-flag}
\end{table*}

\section{Adversarial attack}

\begin{figure}[h]
  \centering
  \includegraphics[width=1\linewidth]{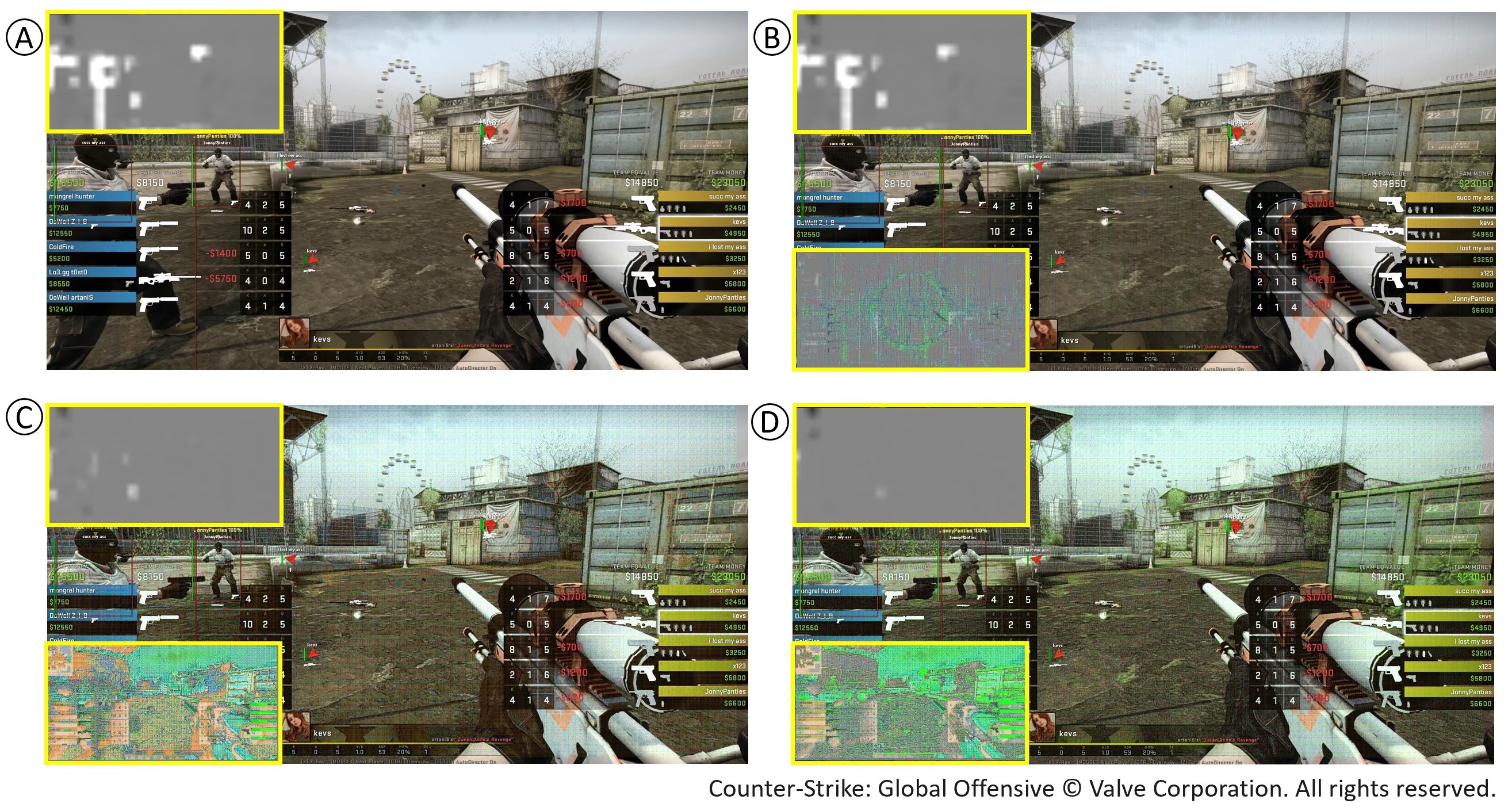}
  \vspace{-1em}
  \caption{Attack strategies: universal attacks (panel B), FGSM (panel C), and Madry (panel D). Corresponding adversarial perturbations and local detection maps are shown in the top/bottom insets. The maximum $L_{\infty}$ norm of the adversarial attacks is $\epsilon=0.1$ (for an image normalized between 0 and 1). \ADD{All the attacks preserve the visual cheating information, but as the strength of the attack increases (Universal $<$ FGSM $<$ Madry), top insets shows that a stronger iterative approach can reduce the visual contrast of essential cheating features, like the red triangles and rectangles in this case.} DNN detector results: Panel A: confident true positive, Panel B: confident true positive, Panel C: non-confident true positive, Panel D: non-confident false negative.}
  \Description{}
  \label{fig:epsilon_0.1_adv_images}
\end{figure} 

\AJ{Adversarial attacks are ad-hoc crafted perturbations, generally imperceptible at visual inspection, that lead the DNN to generate an incorrect output.
In the context of cheating detection, a sophisticated cheating software may fabricate a perturbation aimed at fooling our detector, particularly one that makes the visual hack going undetected.}
In the context of cheating detection, a sophisticated cheating software may fabricate a perturbation aimed at fooling our detector.
Among the several existing attacking strategies, we consider only those that can be practically implemented while interacting with a videogame, i.e., \AJ{those with a small computational cost that can be computed in real-time by the attacking software. }attacking software with a small real-time computational cost.
We also put ourselves in the worst-case scenario where the attacker performs a white-box attack, i.e., has access to the DNN architecture and weights. However, in practice, she may have only access to a proxy DNN and leverage the transferability property of he adversarial attacks to fool our DNN detector.
For these reasons, we limit our investigation to three classes of adversarial attacks, that are, from the less to the most computational\ADD{ly} demanding and effective: universal attacks~\cite{7780651,8099500}, i.e., attacks that are computed offline and can be applied to an entire set of frames; FGSM~\cite{papernot2018cleverhans}, which moves the input image into the direction of the sign of the gradient with respect to the classification cost function, and thus requires only one DNN backward pass to be computed; and PGD (Madry), which iteratively applies FGSM. 

To defend our DNN detector, we use the IBP approach~\cite{DBLP:journals/corr/abs-1810-12715}.
The idea behind the method is simple: given a perturbation with a maximum infinite norm equal to $\epsilon$ in input, one computes a bound for the maximum perturbation in output and train the DNN on the worst-case scenario.
For instance, given the logit vector and the corresponding interval bound propagated on each of its component $(l_0 \pm \delta l_0, l_1 \pm \delta l_1)$, in the case of class 0 being the correct class, would lead to the worst-case scenario $l_0' = l_0 - \delta l_0$, and $l_1' = l_1 + \delta l_1$.
Any cost function (including cross-entropy, MSE, or even the uncertainty loss~\cite{NIPS2018_7580}) can then be computed and minimized on the worst-case scenario, leading to the development of a DNN which is significantly robust to adversarial attacks.

For the \AJ{(simple, indeed) }equations that describe the interval propagation through the linear and non-linear layers of a DNN, we \REM{remand}\ADD{refer} to~
\cite{DBLP:journals/corr/abs-1810-12715}\AJ{. As in the original IBP paper, training stability is achieved by using a cost function that is a linear mix of the original, unprotected cost function and the worst-case cost computed through IBP.
Original IBP paper }, which achieved training stability by using a cost function that is a linear mix of the unprotected original cost function and the worst-case cost computed through IBP.
This helps especially in the beginning of the training process when the intervals are large, and the worst-case scenario output is extremely noisy.
Similarly, we employ curriculum learning to train our DNN detector with IBP by gradually increasing the epsilon value and the IBP loss weight.
We use $1000$ epochs with a batch size of $6$, using Adam optimizer and dropout of $0.15$. The initial learning rate was set at $1e-4$. 
Unlike the \ADD{original} IBP paper, however, we use IBP in combination with the $UL_{global}$ loss on the global head of the DNN detector. \REM{To the best of our knowledge, this is the first time that uncertain loss and IBP are used together.}\ADD{This is the first time the uncertain loss and IBP are used together to the best of our knowledge.}
We train at the same time the local head detector by minimization of $MSE_{local}$ and applying IBP also for the training of the local head.
For the first $150$ epochs, the weight for the KL-divergence component in $UL_{global}$ is gradually increased from 0 to 1, with a learning rate of $1e-4$.
For the next $100$ epochs starts to decay, whereas from epoch 200 to 300, we gradually increase the weight given to the IBP loss, from 0 to 0.5, finally equal the weight of the non-IBP loss.
The $\epsilon$ value remains constant at $1e-8$ until epoch 250, as we noticed that a period with small $\epsilon$ with slowly changing IBP weight adds much-needed stability to the training process;
$\epsilon$ is then gradually increased from $1e-8$ to $0.025$ over the next $250$ epochs, during which the learning rate is $1e-4$.
Finally, we allow the network to stabilize during the fine-tuning phase, with all hyperparameters kept constant.

\ADD{Differently from the general adversarial attack problem widely studied in the literature, where all the classification errors have the same importance, the one considered here is peculiar because of its asymmetry.}
The potential attacker has no practical interest in making a cheat-free frame being detected as a cheating one. \ADD{Even if they try to attack the system by modifying cheat-free frames to fool the network, they will incur a double disadvantage: if cheat-free but modified frames go undetected, the cheater has no practical advantage; on the other hand, recognizing a cheat-free (but modified) frame as a cheating one can be seen as a correct recognition of an artificially modified frame.} Thus, we leverage the problem asymmetry to find a better compromise between clean accuracy and the level of protection.
\ADD{ It is well known in fact that a higher level of protection against adversarial attacks may lead to a decrease in the clean accuracy.} 
Since we only need to protect against attacks that make a cheating frame look like a non-cheating one, we define one-sided IBP, where we consider the worst-case DNN output only for cheating frames. We indicate the resulting loss as $IBP_{One-sided}$ and the original IBP loss as $IBP_{Two-sides}$. 

\subsection{Results}

\begin{figure}[h]
  \centering
  \includegraphics[width=0.9\linewidth]{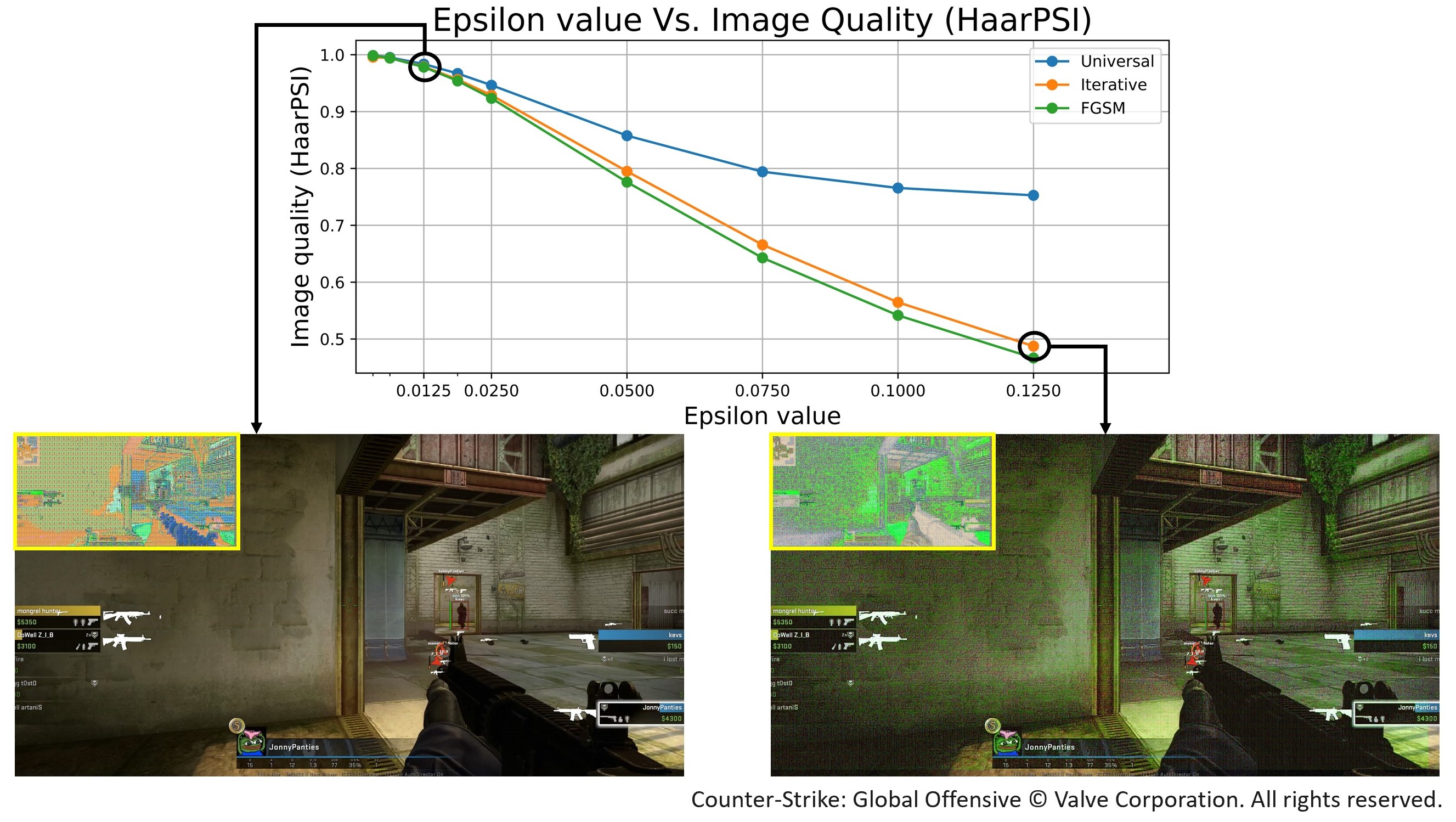}
  \vspace{-1em}
  \caption{Effect of $\epsilon$ on perceived image quality, measured using HaarPSI. $\epsilon$ for left and right images are $0.0125$ and $0.125$.}
  %%\Description{}
  \label{fig:Image_quality_figure}
\end{figure}

To generate adversarial attacks and test the performance of our DNN detector, we use DeepFool~\cite{7780651} to generate universal attack~\cite{8099500} from the entire set of training frames offline, while the FGSM and iterative (Madry) attacks are computed on the fly using Cleverhans ~\cite{papernot2018cleverhans}. We generate only attacks relevant to the anti-cheating problem, i.e., that turn the cheating frame into a non-cheating frame for the DNN detector.
The magnitude of the attacks varies from $\epsilon=0$ to $\epsilon=0.1$, for images normalized between 0 and 1. Notice that an $\epsilon=0.1$ attack is considered extremely strong in the context of adversarial attacks, and defending against such an attack is nowadays considered at least very hard in many practical multi-classes applications.
Some adversarial images for $\epsilon=0.1$ and universal, FGSM and Madry attack are shown in Fig.~\ref{fig:epsilon_0.1_adv_images}.
It is important noticing that even if larger $\epsilon$ values lead to successful attacks, they also corrupt the image potentially up to the point where the gaming experience is ruined (see, for instance, panel D in Fig.~\ref{fig:epsilon_0.1_adv_images}).
Fig.~\ref{fig:Image_quality_figure} shows the effect of $\epsilon$ on image quality. We measure the degree of degradation of a frame corrupted by an adversarial attack through the Haar Wavelet-Based Perceptual Similarity Index (HaarPSI)~\cite{HaarPSI, piq}, which quantifies the perceptual quality\AJ{ to the eye of a human observer, and thus that perceived by the gamer}. HaarPSI value is computed for the adversarial image against an uncorrupted original image. As expected, the image quality decays significantly with an increase in $\epsilon$. 
Therefore, a potential attacker must make a trade-off between the frame quality and the attack success rate.

\begin{figure}[h]
  \centering
  \includegraphics[scale=0.55]{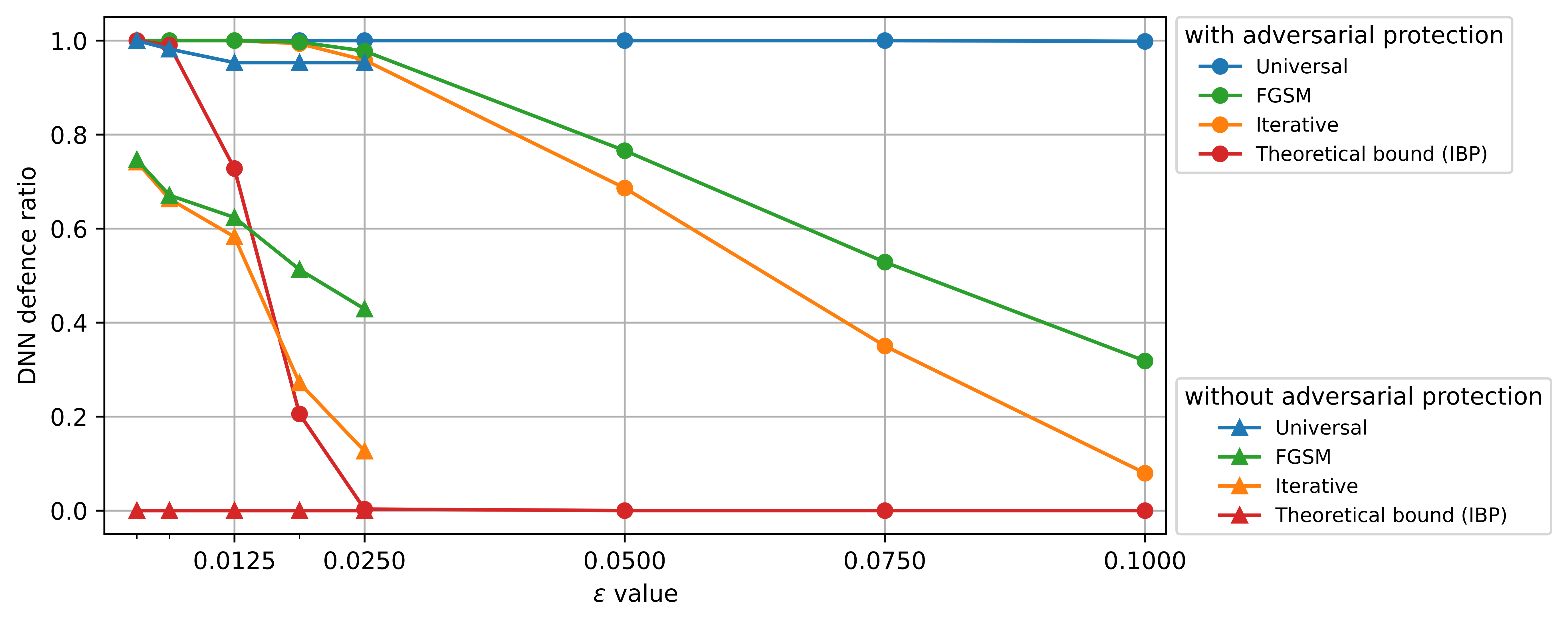}
  \vspace{-1em}
  \caption{Effect of $\epsilon$ on DNN defense ratio $\frac{TP_{\text{attack}}}{TP}$ under different adversarial attacks (universal, FGSM, and iterative) for two networks, with loss functions $MSE_{combined}$ (and thus without any form of adversarial protection) and $UL_{combined}$ + $IBP_{One-sided}$ (and thus trained to be robust against adversarial attacks). For each $\epsilon$ value, we report the least DNN defence ratio among all attacks from $0 \xrightarrow[]{} \epsilon$. We do not consider frames that are classified with low confidence.}
  %%\Description{}
  \label{fig:DNN_accuracy_figure}
\end{figure}

Since we use the confidence analysis introduced in the previous section to determine if a cheating frame should be reported or not, we are interested in measuring how many of these true positives (and confident) are preserved under attack. 
Let then $TP$ and $TP_{\text{attack}}$ be the number of true positives on original frames and frames under adversarial attack.
Since the effectiveness of the attack is measured by how many true positives are converted to false negatives by the attack, we use the ratio of $\frac{TP_{attack}}{TP}$ as a representative of the ability of the network to defend itself during an adversarial attack.
In Fig.~\ref{fig:DNN_accuracy_figure}, we compare the $\frac{TP_{attack}}{TP}$ ratio for the DNN detector trained on the $MSE_{combined}$ loss function and thus without any form of protection against adversarial attack, with that achieved by the detector trained with the $UL_{combined} + IBP_{One-sided}$ cost function, which achieves protection by the combined effect of the uncertainty loss and IBP and guarantees the highest clean accuracy thanks to the asymmetric implementation of the IBP cost function.
This figure demonstrates the importance of protecting the DNN detector against adversarial attacks, as a vanilla implementation of the detector is easily fooled even by small magnitude attacks. In contrast, the ratio of true positive preserved by the protected DNN allows it to operate as an effective cheating detector even under attack, and for large values of $\epsilon$ that already lead to the introduction of significant distortion in the frame (compare with Fig.~\ref{fig:Image_quality_figure}).

\begin{table*}
  \caption{\REM{Effect of IBP \cite{DBLP:journals/corr/abs-1810-12715} and uncertainty loss~\cite{NIPS2018_7580} on TP,  FN, FP, and TN generated by our DNN detector. Non-confident section reports the number of non-confident frames based on the uncertainty criterion, $VR[x]>0$.}\ADD{Effect of IBP \cite{DBLP:journals/corr/abs-1810-12715} and uncertainty loss~\cite{NIPS2018_7580} on the number of true positives (TP, cheating frames reported as cheating), false negative (FN, missed cheating frames), false positives (FP, non-cheating frames incorrectly reported as containing cheating information) and true negative (TN), generated by our DNN detector. The Non-confident section reports the number of frames in each class that cannot be considered legit output because of the low confidence level as determined by the $VR[x]>0$ uncertainty criterion.}}
  \label{tab:Final_performance}
  \vspace{-1em}
  \footnotesize{
  \resizebox{\columnwidth}{!}{
  \begin{tabular}{ccccccccccccl}
    \toprule
    & \multicolumn{4}{c}{\textbf{Confident}} & \multicolumn{4}{c}{\textbf{Non-confident}} & \multicolumn{4}{c}{\textbf{Adversarial attack} ($\epsilon$ = 0.0125)}  \\ \cmidrule(r){2-5} \cmidrule(r){6-9} \cmidrule(r){10-13} 
    Loss function & TP & FN & FP & TN & TP & FN & FP & TN & Universal & FGSM & Iterative & IBP bound\\
    \midrule
    $MSE_{combined}$ & 389 & 1030 & 57 & 1354 & 45 & 26 & 39 & 40 & 0.964 & 0.6221 & 0.5809 & -\\
    \midrule
    $UL_{combined}$ & 525 & 606 & - & 1429 & 145 & 214 & 2 & 59 & 0.979 & 0.8057 & 0.7752 & -\\
    $IBP_{Two-sides}$ & 447 & 764 & 132 & 1102 & 203 & 76 & 163 & 93 & 1.0 & 0.6464 & 0.6375 & 0.550\\
    $IBP_{One-sided}$ & 353 & 873 & 53 & 1172 & 189 & 75 & 173 & 92 & 1.0 & 0.745 & 0.7422 & 0.6288\\
    \midrule
    $UL_{combined}$ + $IBP_{Two-sided}$ & 591 & - & 8 & - & 275 & 624 & 60 & 1422 & 1.0 & 1.0 & 0.9983 & 0.6531\\
    $UL_{combined}$ + $IBP_{One-sided}$ & 628 & - & 1 & - & 284 & 578 & 14 & 1475 & 1.0 & 1.0 & 1.0 & 0.7277\\
  \bottomrule
\end{tabular}}}
\end{table*}

To better understand the contribution of the loss function (and in particular of $UL_{global}$) and IBP and how they interact when the DNN is under an adversarial attack, we performed an ablation study and measured the TP, FN, FP, and TN statistics for the same DNN architecture trained with the different cost functions.
For the ablation study, we set $\epsilon=0.0125$, which results in a reasonable trade-off between the attack's strength and the frame quality, with a theoretical bound for defense provided by IBP above $50\%$.
In all the cases, the $VR[x]>0$ criterion is applied to reject low confidence detections. Besides, we discharge frames with high uncertainty value $u$ as well.
Results are reported in Table~\ref{tab:Final_performance}, where the first two rows show that, when IBP is not used in training, the theoretical protection bound is zero, suggesting that an attack is potentially possible on the entire dataset.
The practically feasible attacks considered here (universal, FGSM, Madry) can indeed decrease the detector's accuracy to $58\%$ in the worst case. In this situation, our DNN can still find many (389 TP) of the existing cheating frames, but this situation is not desirable as there is no theoretical robustness guarantee.
The lack of a theoretical guarantee bound also affects the DNN trained with the $UL_{combined}$ loss (second row), which on the other hand, provides higher accuracy against the practical attacks, as already noticed in~\cite{NIPS2018_7580}.
The adoption of IBP without the uncertainty loss (third and fourth rows) increases the practical robustness of the DNN with respect to the vanilla $MSE_{combined}$ training, and an IBP bound can now be established to guarantee that more than $55\%$ of the frames in the dataset cannot be attacked.
However, a side effect of IBP training is to increase the number of false positives (132) on clean data, a type of error that is incompatible with the practical deployment of an effective anti-cheating system.
Training with one-side IBP reduces FP, but does not entirely solve the problem.
The best results are achieved when both uncertainty loss and IBP are used: FP is minimized, and almost full protection against practical adversarial attacks is achieved simultaneously. 
Also, in this case, leveraging the problem's symmetry with one-side IBP leads to the best compromise in terms of clean accuracy and level of robustness for the class of interest, with only one FP reported on the entire dataset, and 628 TP.

Overall, the results presented in these last sections demonstrate that the development of a reliable and precise enough anti-cheating detector requires a careful design that considers several aspects at the same time.\AJ{, from the DNN architecture that guarantees that the attention of the detector is not focused on the main cheating features only, to the estimate of the level of confidence that is used to avoid reporting legit players as cheaters and to determine the need for DNN re-training when new games or cheating software are introduced on the market.
Considering the possibility that the cheating software may use adversarial attacks to attack the DNN detector makes its development even more complicated. However,} We have demonstrated that the adoption of state of the art techniques for confidence estimation and adversarial defense, together with the leverage of the peculiar aspects of the cheating detection problem, allows creating a system that meets the practical requirements for its deployment on the field.
\section{Conclusion}
We presented a comprehensive system for hack detection from rendered visual information that identifies cheating without risking operating system security or privacy concerns.
Crucially for deployment, our paper anticipates adversarial attacks and shows that the combined effect of uncertainty loss and one-sided IBP leads to a DNN capable of detecting cheating frames, both on clean data and under attack, while minimizing the false positives. 
For reproducibility and followup work, we released our dataset.%\ADD{~\cite{public_dataset}}\REM{at \textit{hidden URL}}.
\ADD{Although our DNN approach is robust to the evolution of legitimate game graphics and cheating software, it is not invulnerable. At some point, the prior visual distribution of games or cheating features will shift.}
By comparing the likelihood ratios measured in training with those achieved once deployed on the field, we demonstrated that we could automatically identify situations where the DNN detector is not operating with the expected confidence and requires re-training.
\AJ{Validating whether a user has legitimately satisfied a testing challenge in a virtual environment is a problem not limited to games. Nor is ensuring that such validation preserves user privacy and computer security. We observe that the educational online testing community has historically struggled with both parts of this, and the present global pandemic has certainly highlighted this problem. Desktop and virtual reality 3D environments for corporate compliance, emergency responder training, and physical tasks are very similar to our considered use case of games and are natural next step for improved anti-cheat techniques. Moreover, natural video, text-only, or audio-only computing environments are relevant and important domains for this technology.}
\ADD{Since the input to the model is made of individual frames, our model is robust against cheats applied at irregular time intervals. By using techniques like optical flow, our work can be extended to game videos.
Data from keyboard inputs, game logs, and network traffic can be combined with visual input to create a robust cheat-detection system.}
Generalizing from our result, we hypothesize that a DNN approach using only externally visible I/O data and executing on a local computer or peripheral will be an effective strategy for more anti-cheating problems.

%% Uncomment for including acknowledgements
%%\begin{acks}
%%\end{acks}

\bibliographystyle{ACM-Reference-Format}
\bibliography{references}

\end{document}